\documentclass{article}

\usepackage{arxiv}

\usepackage[utf8]{inputenc}
\usepackage[T1]{fontenc}
\usepackage{hyperref}
\usepackage{url}
\usepackage{booktabs}
\usepackage{amsmath, amssymb, amsfonts}
\usepackage{nicefrac}
\usepackage{microtype}
\usepackage{graphicx}
\usepackage{natbib}
\usepackage{doi}

\usepackage{float}
\usepackage{multirow}   

\usepackage{subcaption} 
\usepackage{infwarerr}
\title{Assignment-Routing Optimization: Solvers for Problems under Constraints}
\author{
  {\includegraphics[scale=0.06]{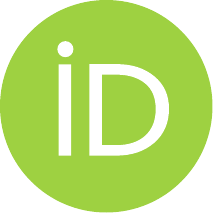}\hspace{1mm} Yuan Qilong $^{*a}$,  Michal Pavelka$^{b}$. } \\$^{a}$ Singapore Institute of Technology. $^{b}$ Mathematical Institute, Faculty of Mathematics and Physics, Charles University
\\  \texttt{$^*$ Corresponding Author Email: qilong.yuan@singaporetech.edu.sg} \\
}

\hypersetup{
  pdftitle={},
  pdfsubject={Optimization, Combinatorics, Gurobi},
  pdfauthor={Yuan Qilong, Michal Pavelka},
  pdfkeywords={Joint Routing-Assignment, Time-Window Constraints, 
Mixed-Integer Programming, 
Combinatorial Optimization, Multi-Type Object Packaging Planning},
}

\begin{document}
\maketitle

\begin{abstract}
We study the Joint Routing-Assignment (JRA) problem in which items must be assigned one-to-one to placeholders while simultaneously determining a Hamiltonian cycle visiting all nodes exactly once. Extending previous exact MIP solvers with Gurobi and cutting-plane subtour elimination, we develop a solver tailored for practical packaging-planning scenarios with richer constraints. These include multiple placeholder options, time-frame restrictions, and multi-class item packaging. Experiments on 46 mobile manipulation datasets demonstrate that the proposed MIP approach achieves global optimality with stable and low computation times, significantly outperforming the  shaking-based exact solver by up to an orders of magnitude. Compared to greedy baselines, the MIP solutions achieve consistent optimal distances with an average deviation of 14\% for simple heuristics, confirming both efficiency and solution quality. The results highlight the practical applicability of MIP-based JRA optimization for robotic packaging, motion planning, and complex logistics . GitHub repository:\url{https://github.com/QL-YUAN/JAR_Time_Frame_And_Additional_Phs_Constraints.git}
\end{abstract}

\section{Introduction}
\label{sec:intro}


In this paper, we extend the JRA framework by developing a Joint Assignment-Routing (JAR) solver that supports:
\begin{itemize}
  \item Time-frame constraints: items and placeholders can be active only during certain intervals, and items may choose among multiple feasible time windows.
  \item Multiple item and placeholder types: enabling packaging scenarios where different classes of objects must be grouped or allocated to different placeholder types.
  \item Additional Placeholders: items have more placeholders options for a more optimal allocation arrangement.
\end{itemize}

To handle these added complexities, we propose solver variants that build upon MIP solvers, adapted to incorporate additional placeholder, time-window and type constraints \cite{kara2004note,basu2020complexitybranchandboundcuttingplanes,gurobi2025mipprimer}. We demonstrate through experiments on synthetic and realistic problem instances that our solver is both efficient (significantly faster than baseline methods) and effective (achieving global optima under complex constraints).
We present a detailed empirical evaluation, showing how time-frame conditional methods can yield large runtime improvements (e.g., up to one orders of magnitude faster) while maintaining solution quality, and discuss implications for robotics, motion planning, and logistics.

By providing both a benchmark dataset and solver implementations, we hope this work serves as a foundation for future research on scalable algorithms for joint routing-assignment and packaging planning under realistic constraints.

\section{Mathematical Formulation}
\subsection{Joint Routing-Assignment (JRA) Problem General Formulation}
Start with $n$ items to be allocated to $n_p$ placeholders. This section discusses the one-to-one item-placeholder allocation problem. Then, $n_p=n$.
Let $I = \{0, \dots, n-1\}$ be the set of items, $P = \{n, \dots, 2n-1\}$ the set of placeholders, and $c_{ij}$ the cost of traversing edge $(i,j)$. Denote $V = I \cup P$ and $E = \{(i,j) \in V \times V : i \neq j\}$. 

We define binary decision variables:
\[
x_{ij} = \begin{cases} 1 & \text{if edge } (i,j) \text{ is in the cycle} \\ 0 & \text{otherwise} \end{cases}, 
\quad
a_{ip} = \begin{cases} 1 & \text{if item } i \text{ is assigned to placeholder } p \\ 0 & \text{otherwise} \end{cases}.
\]

The objective is to minimize total traversal cost:
\[
\min \sum_{(i,j) \in E} c_{ij} x_{ij}.
\]

Constraints are as follows:

    \textbf{g1. Degree:} Each node has exactly one input and one output incident edges:
    \[
    \sum_{j \in V, j \neq i} x_{ij}=1,  \sum_{j \in V, j \neq i} x_{ji} = 1, \quad \forall i \in V
    \]
    \textbf{g2. Assignment:} Each item and placeholder is assigned twice, whereby links are always from item to placeholder:
    \[
    \sum_{p \in P} a_{ip} = 2, \quad \forall i \in I, 
    \quad
    \sum_{i \in I} a_{ip} = 2, \quad \forall p \in P
    \]
    \textbf{g3. Valid edges:} Only item-to-placeholder edges are allowed, and edges can only exist if the assignment exists:
    \[
    x_{ij} = 0 \text{ if both } i,j \in I \text{ or } i,j \in P, 
    \quad
    x_{ip} \le a_{ip}, \quad x_{pi} \le a_{ip}, \forall i \in I, p \in P
    \]
    \textbf{g4. Fixed pair:} In case of having starting and goal points, we specify starting position to be the last placeholder, goal point to be the last item, and enforce the last item to be assigned to the last placeholder as following constraints. 
\[
a_{n-1, 2n-1} = 1
\]

\[
x_{n-1, 2n-1} = 1,x_{2n-1, n-1} = 0
\]
\textbf{g5. Subtour elimination:} Disconnected cycles are prohibited by a cutting-plane method \cite{kelley1960cutting,westerlund1995extended}:
\[
\sum_{i,j \in S, i \neq j} x_{ij} \le |S|-1, \quad \forall \text{ subtours } S \subset V
\]
This is enforced dynamically using subtour elimination callback function in Gurobi. Another alternative way of Sub-tour Elimination is through applying as introduced in next sub-section.

Beyond standard form of JRA problem, there are variation for transforms. The variation with additional placeholder options, joint packaging of multiple classes of objects are discussed in Appendix with demonstration examples. Next sub-section focus on formulation to solve JRA problem under time-frame restrictions.

\subsection{JRA problem with Time Frame Section Constraints}

In many practical applications, items may be subject to \emph{time frame or section constraints}, meaning that each sub-group of items can only be allocated within a designated stage or time window. Formally, suppose there are $h$ time frames, and let 
\[
G_m \subseteq I, \quad m = 1, \dots, h
\] 
denote the set of items belonging to time frame $m$. The items must be handled sequentially according to their time frame, i.e., handing sequentially $G_1$, $G_2$, and so on until $G_h$.

\subsubsection{Item Graph Formulation with Time-Frame Constraints}

\medskip

To model the sequential processing of items across multiple time frames, we construct a directed \emph{item graph} over the set of items \(I=\{1,\dots,n\}\). For every ordered pair of distinct items \((i,j)\), we define binary variables
\[
y_{ij} =
\begin{cases}
1, & \text{if item $j$ is handled immediately after item $i$},\\[3pt]
0, & \text{otherwise},
\end{cases}
\qquad i,j\in I,\; i\neq j.
\]

Additionally, each item \(i\in I\) is assigned an integer sequencing variable
\[
t_i \in \{1,\dots,n\},
\]
indicating the position of item \(i\) in the global sequence.

\noindent
\textbf{C1. Section-Based Temporal Precedence}

The items are partitioned into \(h\) ordered sections (time frames),$G_1,\, G_2,\, \dots,\, G_h$, where section \(G_m\) must be processed before section \(G_{m+1}\). To enforce this, we impose
\[
t_k \;\ge\; t_i + 1,
\qquad \forall\, i\in G_m,\; k\in G_{m+1},\quad m=1,\dots,h-1.
\]
This guarantees a strictly increasing sequence across sections and prohibits interleaving of items from different time frames.

\noindent
\textbf{C2. Special Constraints for the Stop Item \(P_n\)}

\(P_n\) denote the item representing Stop ( stop is defined as an item and assigned to the last item), and \(P_n=|I|=n\). 
Item \(P_n\) must appear no earlier than any item in the last section \(G_h\):
\[
t_{P_n} \;\ge\; t_i,
\qquad \forall\, i\in G_h.
\]

\noindent
\textbf{C3. Subtour Elimination ( Constraints)}

To ensure that the item graph forms a single continuous sequence without subtours, we apply Miller-Tucker-Zemlin constraints \cite{kara2004note}. For all distinct items \(u,v\in I\), with \(u\neq P_n\),
\[
t_u - t_v + M\, y_{uv} \;\le\; M - 1,
\qquad M = n.
\]
If \(y_{uv}=1\), this forces \(t_v \ge t_u + 1\); otherwise, the constraint is non-binding. These constraints prevent the formation of subtours and ensure a single valid sequence consistent with the section precedence rules.

\medskip
\noindent
\textbf{C4. Total edges sum:}  
The total number of edges in the item graph equals the total number of items $|I|$:
\[
\sum_{i \in I} \sum_{j \in I, j \neq i} y_{ij} = |I|.
\]

\medskip
\noindent
\textbf{C5. Single successor per item:}  
Each item is followed by exactly one other item:
\[
\sum_{j \in I, j \neq i} y_{ij} = 1, \quad \forall i \in I.
\]

\medskip
\noindent
\textbf{C6. Single predecessor per item:}  
Each item is preceded by exactly one other item:
\[
\sum_{j \in I, j \neq i} y_{ji} = 1, \quad \forall i \in I.
\]

\medskip
\noindent
\textbf{C7. Optional symmetry-breaking:}  
To accelerate computation, constraints can be added to forbid bidirectional edges between any pair:
\[
y_{ij} + y_{ji} \le 1, \quad \forall i,j \in I, \, i \neq j.
\]

\medskip
\noindent
\textbf{C8. One placeholder linking two items in between:} Once the item graph constraints are confirmed, placeholders need to be assigned to items with respect to such constraints. Specifically, placeholders can only be placed along the connections defined by $y_{ij}$. Let $x_{ip}$ indicate assignment of placeholder $p$ to item $i$, and $x_{pj}$ indicate the subsequent placeholder connection to item $j$. Then, the following must hold:
\begin{equation}
\sum_{p \in P} x_{ip} \, x_{pj} = y_{ij}, \quad \forall i,j \in I, \, i \neq j.
\label{eq:yij_definition}
\end{equation}

Constraints from C1 to C8 together with the general JRA formulation forms complete solver to solve the JRA problem under Time-Frame Constraints.

\subsubsection{Subtour Elimination via MTZ Constraints on Item Graph}

In the original formulation as introduced in \cite{yuan2025datasets}, subtours- disconnected cycles among items-were prevented indirectly through subtour elimination constraints applied to the routing variables $x_{ij}$. With the introduction of explicit Miller-Tucker-Zemlin (MTZ) constraints on the item graph variables $y_{ij}$, such indirect enforcement is no longer necessary.

The MTZ constraints explicitly enforce a valid ordering of items along any path defined by $y_{ij}$, thereby eliminating cycles among items at the $y$-level.

Crucially, in the proposed model, every edge $y_{ij}=1$ implies the existence of a corresponding path through placeholder nodes in the routing graph $x_{ip}, x_{pj}$. 

Formally,
\[
y_{ij} = 1 \implies \exists \, p \in P: x_{ip} \, x_{pj} = 1,
\]
where $P$ denotes the set of placeholder nodes. Therefore, any subtour that would form among items under $y$ would automatically induce a corresponding cycle in $x$. Since the MTZ constraints on $y$ prevent cycles at the item level, the corresponding $x$-level subtours cannot occur. As a result, the original subtour elimination constraints for $x_{ij}$ are redundant and can be safely removed.

\medskip
\noindent
\textbf{Remarks.}  
By enforcing MTZ constraints on $y_{ij}$, the model guarantees a sequential, group-wise routing of items while inherently preventing cycles in both the item and placeholder graphs. This simplification reduces solver complexity, eliminates the need for subtour elimination callbacks on $x_{ij}$, and preserves the temporal and group-based consistency of item handling paths. This is particularly advantageous in scheduling, assembly, and robotic manipulation applications with sequencing requirements, especially when Time-Frame constraints are involved.

\section{Complexity and Comparison Solvers}
\subsection{Combinatorial Complexity}
\label{sec:combinatorial}
Joint robotic assignment (JRA) problems under time-frame constraints exhibit extremely high combinatorial complexity. Consider the following illustrative scenario:

\begin{itemize}
    \item 16 items are to be collected.
    \item The mobile manipulator visits 4 stops.
    \item The distribution of items per stop is: 2, 5, 5, and 4.
    \item There are 18 optional placeholders for placing these items.
\end{itemize}

The total number of possible pick-and-place combinations can be expressed as:

\[
2! \, P(18,2) \;\times\; 5! \, P(16,5) \;\times\; 5! \, P(11,5) \;\times\; 4! \, P(6,4),
\]

where \(P(n,k) = \frac{n!}{(n-k)!}\) denotes the number of permutations of \(n\) elements taken \(k\) at a time. Evaluating each term:

\[
\begin{aligned}
2! &= 2, & P(18,2) &= 306,\\
5! &= 120, & P(16,5) &= 524{,}160,\\
5! &= 120, & P(11,5) &= 55{,}440,\\
4! &= 24, & P(6,4) &= 360.
\end{aligned}
\]

Multiplying these together yields

\[
2 \times 306 \times 120 \times 524{,}160 \times 120 \times 55{,}440 \times 24 \times 360 \approx 1.82 \times 10^{20}.
\]

This astronomical number highlights the infeasibility of a brute-force solution even for a relatively small instance of the problem, emphasizing the need for efficient heuristic or optimization-based approaches in practical JRA scenarios.

\subsection{Shaking-Based Solver for Time-Frame Constrained JRA}

In previous work \cite{qilong2025assignmentroutingoptimizationefficient,yuan2025efficientoptimalsolverjoint}, we introduced a \emph{shaking algorithm} for addressing general Joint Robotic Assignment (JRA) problems. This algorithm can be directly applied to \textbf{exactly} solve the time-frame constrained JRA problem considered here. However, the computational cost is significant, asrequiring tens of seconds per instance shown results , due to the combinatorial nature of the problem. While heuristic approaches can provide near-optimal solutions much faster, they are not exact and are therefore not discussed in this work.

The shaking-based solver is implemented as follows: under the time-frame constraints, all possible \emph{instants} are generated, where an instant is defined as a complete, unique sequence of items respecting the time-frame ordering. Mathematically, if the items are distributed across \(S\) stops with \(n_1, n_2, \dots, n_S\) items per stop, the total number of instants is:

\[
\prod_{s=1}^{S} n_s!,
\]

where \(n_s!\) accounts for all permutations of items at stop \(s\). For the example in Section~\ref{sec:combinatorial}, with 4 stops having \(2, 5, 5, 4\) items respectively, the total number of instants is

\[
2! \times 5! \times 5! \times 4! = 2 \times 120 \times 120 \times 24 = 691{,}200.
\]

For each instant, a \emph{single-time} shaking algorithm, based on the Hungarian method, which has the computaional complexity of $n^3$ (n=16 in this example), is executed to obtain a local optimal assignment of items to placeholders that satisfies the time-frame constraints. Let \(L(\mathcal{I})\) denote the local optimum for instant \(\mathcal{I}\). The global optimum is then obtained as:

\[
\mathcal{O}^* = \max_{\mathcal{I} \in \text{Instants}} L(\mathcal{I}),
\]

where the maximization is performed over all instants.  

While this procedure guarantees an exact solution, its computational burden grows rapidly with the number of items and stops, reinforcing the combinatorial complexity discussed in Section~\ref{sec:combinatorial}. This solver is implemented and experimentally tested to compare with the introduced MIP optimization solver.

\section{Experiments}
\subsection{Datasets}
The experimental dataset is derived from a mobile manipulation scenario in which a mobile manipulator autonomously visits a sequence of predetermined locations. At each stop, the on-board robotic arm performs a tabletop pick-and-place operation: simple items located on the table surface are grasped and subsequently placed in designated placeholders on the robot. This workflow is conceptually similar to classical robotic automation settings where a fixed-base manipulator interacts with a conveyor system. In such conveyor-based systems, the continuous motion of the conveyor divides the arm's motion into discrete time-frame segments corresponding to different conveyor sections.

In mobile manipulation, the robot base itself is in motion; however, when focusing on task and motion planning for the arm, the relative behavior closely parallels the conveyor-based paradigm. Each stop of the mobile base creates a time-frame section in which a subset of items becomes reachable and is collected.

A total of 46 experimental samples are provided in the file \textit{datasets.pkl}. These samples can be loaded using the provided \textit{data\_loader.py} script available in the associated GitHub repository.

For each sample, the field \textit{sps} contains a list of grouped localized (in robot base frame) 2D item positions. Each group corresponds to the set of item locations collected during a single stationary phase of the mobile base. Thus, the index of the sample denotes the experiment identifier, while the length of the \texttt{sps} list indicates the number of time-frame segments generated by base stops.

Across all experiments, exactly 16 items are collected. The segmentation of these items into groups within \texttt{sps} reflects the temporal and spatial structure imposed by the robot's navigation and stopping behavior, which is in theory much more complex than explained here. However, the focus of this paper is to validate the solver capability for JRA problem under Time-Frame constraint. Therefore, the datasets only need to be considered as inputs for algorithm validation. 

\subsection{Experiment Setup}
\paragraph{Computer Hardware}
All experiments were conducted on a workstation equipped with an Intel(R) Core(TM) Ultra 7 155H processor, supporting the SSE2, AVX, and AVX2 instruction sets. The CPU has 22 physical cores and 22 logical processors. With both algorithm applying parallel computing technology with full CPU count setting.

\paragraph{Computation Experiment}
Both the shaking based algorithm and the proposed MIP optimization algorithm are tested with all the dataset samples. For each sample with the sectioned items to collect and an additional section representing the start/stop of the robot are included. 

\subsection{Computation Experiment}

Both the shaking-based solver and the proposed MIP optimization approach were tested on all dataset samples. Each sample consists of sectioned items to collect, including an additional section representing the start/stop of the robot. A simple greedy approach is also included as a baseline for comparison.

\textbf{MIP Optimization}: Solves the time-frame constrained JRA problem exactly using the proposed Mixed-Integer Programming formulation.

\textbf{Shaking-Based Algorithm}: Enumerates all possible \emph{instants} (unique item sequences per time-frame) and applies a Hungarian-based single-time shaking algorithm to compute local optima. The global optimum is the best among all instants.

\textbf{Greedy Baseline}: A simple heuristic assigning items sequentially to the nearest available placeholders, without guaranteeing optimality.

\subsubsection{Computation Results}
\begin{figure}[h]
  \centering
  \includegraphics[width=0.62\linewidth, trim=3mm 0mm 0mm 0mm, clip]{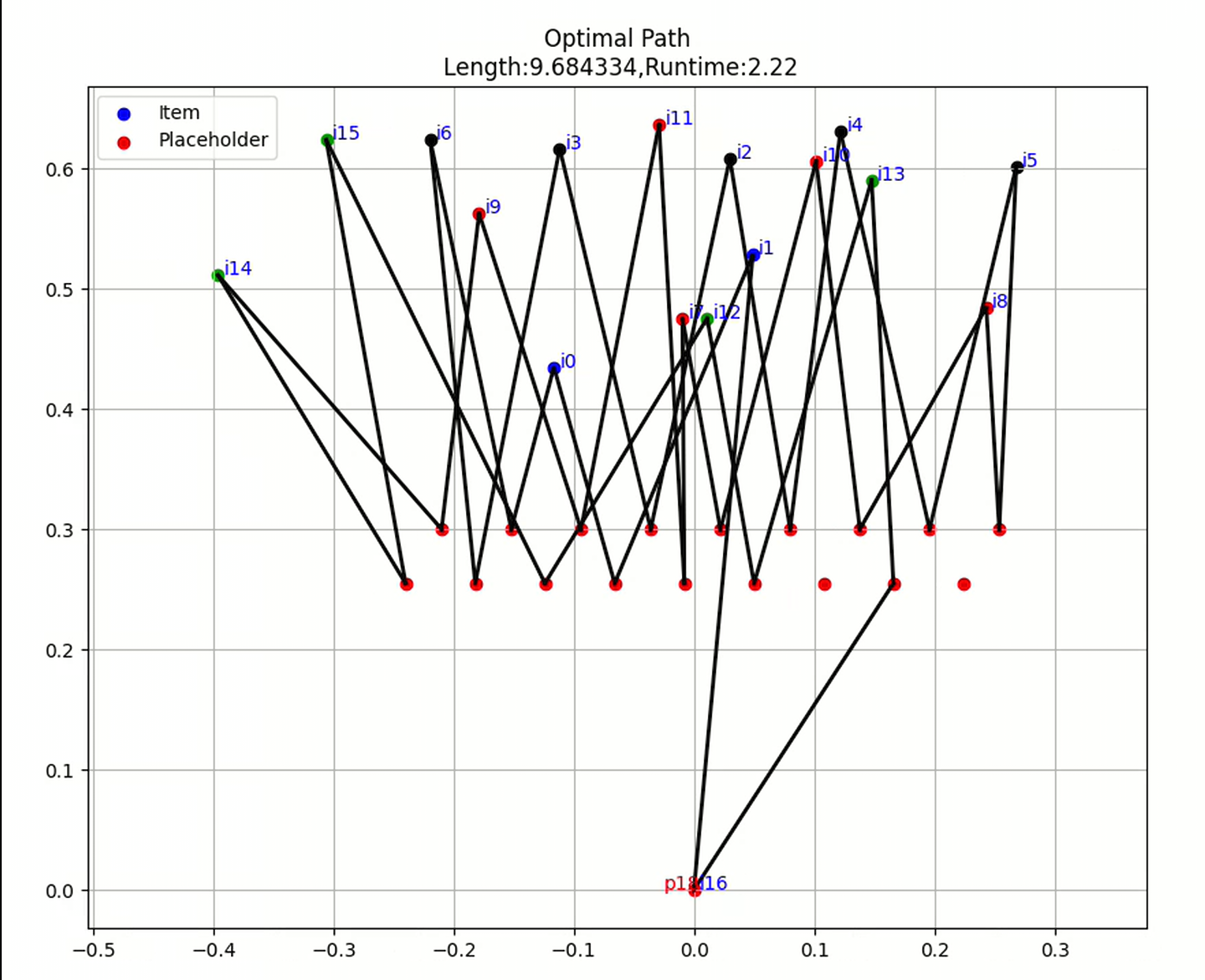}
  \caption{Comparison of computation times between MIP and shaking-based solvers.}
  \label{fig:QMA_S}
  \vspace{0mm}
\end{figure}
Figure \ref{fig:QMA_S} demonstrates a sample of the calculation results with different colors indicating different section of object items (including the start/stop point). As mentioned, all the points presented here are localized results.  

Table~\ref{tab:combined_results_updated} summarizes computation time and total distance for all 46 samples.  
For time, the table includes MIP and Shaking per-sample results with an average row at the end.  
For distance, the table shows MIP (exact), Shaking (exact), and Greedy, along with the percentage difference of Greedy relative to MIP. The average error is also computed at the bottom.
\begin{table}[H]
\centering
\caption{Computation Results with Timing and Distance Comparisons (Updated with Nss)}
\label{tab:combined_results_updated}
\begin{tabular}{c|cc|ccc|c|c}
\hline
Index & \multicolumn{2}{c|}{Time [s]} & \multicolumn{3}{c|}{Distance [m]} & Greedy Error [\%] & Nss \\
\cline{2-8}
& QMA & MIP & QMA & MIP & Greedy & $\Delta = \frac{\text{Greedy}-\text{MIP}}{\text{MIP}} \cdot 100$ & \\
\hline
1 & 65.4953 & 3.5000 & 10.6878 & 10.6878 & 11.5451 & 8.0214 & 518400 \\
2 & 2.9202 & 1.1326 & 10.8181 & 10.8181 & 12.2787 & 13.5015 & 23040 \\
3 & 17.5983 & 3.4517 & 11.0760 & 11.0760 & 12.7978 & 15.5446 & 138240 \\
4 & 5.4688 & 4.2849 & 12.3483 & 12.3484 & 13.9023 & 12.5830 & 41472 \\
5 & 7.2784 & 1.3362 & 10.2573 & 10.2573 & 11.2651 & 9.8253 & 55296 \\
6 & 8.8747 & 1.8113 & 9.7618 & 9.7618 & 10.5510 & 8.0844 & 69120 \\
7 & 3.4117 & 5.3364 & 10.1608 & 10.1608 & 11.5955 & 14.1198 & 25920 \\
8 & 1.8576 & 1.7763 & 10.9467 & 10.9467 & 12.1208 & 10.7260 & 13824 \\
9 & 6.4860 & 4.6426 & 10.3314 & 10.3314 & 11.8677 & 14.8703 & 51840 \\
10 & 4.4699 & 4.3632 & 10.6455 & 10.6455 & 12.2969 & 15.5126 & 34560 \\
... & ... & ... & ... & ... & ... & ... & ... \\
\hline
Average & 34.4606 & 2.9638 & 10.5345 & 10.5345 & 12.0371 & 14.3028 & -- \\
\hline
STDEV & 71.7693 & 1.6088 & 0.7841 & 0.7841 & 0.9003 & 3.1927 & -- \\
\hline
\end{tabular}
\end{table}

\paragraph{Analysis}  
\begin{itemize}
    \item \textbf{Time Efficiency:} The MIP solver exhibits an average computation time of 3.0 s per sample, significantly faster than the shaking-based algorithm with an average of about 35 s. This demonstrates MIP’s practical efficiency for exact solutions. Shaking based algorithm efficiency directly depends on the structure of the item distribution as introduced in Section 3. The QMA solver time is actually propotional to the total sample number, Nss as shown in Table I. MIP solver's efficiency, on the other hand, is very stable with much lower standard deviation.  
    \item \textbf{Accuracy:} Greedy solver is almost immediate, but the results have an average error of 14.0\% relative to the MIP optimum, confirming that simple heuristics are consistently suboptimal.  
    \item \textbf{Shaking vs MIP:} Both achieve identical distances for all samples, proving that MIP produces exact global optima while being computationally far more efficient than the exhaustive shaking-based approach.
\end{itemize}

\paragraph{Summary}  
These results demonstrate that the proposed MIP optimization method provides a favorable balance between \textbf{optimality} and \textbf{computational efficiency}. Compared to the shaking-based algorithm, MIP delivers exact solutions with much lower and more stable runtimes, making it highly suitable for practical tasks and motion planning under time-frame constraints.

In this work, we investigated the Joint Robotic Assignment (JRA) problem under time-frame constraints. A comprehensive dataset of 46 experiments was utilized, where a mobile manipulator collects and places items across multiple stops, emulating practical robotic manipulation scenarios.

We analyzed the combinatorial complexity of the JRA problem, showing that even for a moderate setup with 16 items, 4 stops, and 18 optional placeholders, the number of feasible pick-and-place sequences grows to the order of $10^{20}$. This demonstrates that brute-force solutions are computationally infeasible, motivating the need for efficient optimization techniques.

Two solution strategies were evaluated: a shaking-based solver, capable of finding exact solutions under time-frame constraints, and a Mixed-Integer Programming (MIP) formulation designed to achieve global optimality more efficiently. Experimental results revealed that the MIP approach consistently produced solutions matching the shaking-based method in terms of distance metrics, while significantly reducing computation time. Importantly, the MIP solver demonstrated stable runtimes across all instances, whereas the shaking-based approach was highly sensitive to the distribution of items and stops.

Quantitative analysis across all samples showed that the MIP solver reduced computation times by orders of magnitude compared to the shaking-based approach, while maintaining high accuracy, with greedy solutions deviating on average by 14.0\% from the optimal paths. These results underscore the practicality of the proposed MIP approach for real-time or near real-time mobile manipulation tasks involving complex assignment and routing challenges.

In conclusion, the MIP-based optimization method offers a robust and computationally efficient framework for solving time-frame constrained JRA problems. Future work will explore strategies for complete robot task and motion planning with such MIP solver being applied.

\section*{Appendix}

\subsection*{A. Extension to Problem with More Placeholder Options}

The formulation presented in the Section 2.1 addresses the case where the number of items and placeholders are equal, i.e., \( n_p = n \). In this configuration, each item is assigned to exactly one placeholder, and each placeholder accommodates exactly one item. However, in many practical scenarios, the number of available placeholders may exceed the number of items (\( n_p > n \)), offering additional flexibility in the allocation process.

\noindent
When \( n_p < n \), the problem becomes infeasible for a single traversal cycle since not all items can be allocated. Therefore, such cases are not considered further in this work. The following discussion focuses on the scenario where \( n_p > n \), meaning there exist more potential placeholders than items.

\subsubsection*{Complete Formulation with Placeholder Selection Variables.}

One possible approach to handle the case of \( n_p > n \) is to introduce additional binary selection variables:
\[
c_p = 
\begin{cases}
1, & \text{if placeholder } p \text{ is selected for use}, \\
0, & \text{otherwise},
\end{cases}
\quad \forall p \in P,
\]
and impose the constraint:
\[
\sum_{p \in P} c_p = n,
\]
to ensure that exactly \( n \) placeholders are selected.

\textbf{Degree:} Updates need to be applied to incident edges. 
For items:
    \[
    \sum_{j \in V, j \neq i} x_{ij}=1,  \sum_{j \in V, j \neq i} x_{ji} = 1, \quad \forall i \in V
    \]
For placeholders:
\[
\sum_{j \in V, \, j \neq p} x_{pj}=\, c_p,  \sum_{j \in V, \, j \neq p} x_{jp} = \, c_p, 
\quad \forall p \in P,
\]
where \( c_p \in \{0,1\} \) indicates whether placeholder \( p \) is selected in the solution.

\textbf{Assignment:} Each item and placeholder assignment depends on the selected placeholders. The assignment constraints become:
    \[
    \sum_{p \in P} a_{ip} = 2, \quad \forall i \in I,
    \qquad
    \sum_{i \in I} a_{ip} = 2c_p, \quad \forall p \in P,
    \]
These ensure that each item is assigned to exactly one placeholder, and only the selected placeholders can be assigned to items. 

The assignment variable \( a_{ip} \) is then coupled with the placeholder selection variable:
\[
a_{ip} \le c_p, \quad \forall i \in I, \; p \in P.
\]
This ensures that items can only be assigned to selected placeholders. The problem become optimizing the same cost function, subject to all previous assignment, degree, and routing constraints, along with the additional placeholder selection constraints above.

\medskip

\subsection*{B. Extension to Multiple Item Types}

Many practical applications involve multiple types or categories of items, each associated with a distinct subset of compatible placeholders. Examples include multi-type packing tasks, board game piece arrangements, and robotic placement operations where items of the same type are interchangeable but must be placed in designated regions.

To generalize the formulation for such cases, we partition the sets of items and placeholders according to their types. Let the index set of item types be \( S_p = \{1, \dots, |S_p|\} \). For each type \( s_p \in S_p \), denote:
\[
I_{s_p} \subseteq I \quad \text{and} \quad P_{s_p} \subseteq P
\]
as the subsets of items and placeholders of type \( s_p \), respectively. Here we assume that this problem is already a one-to-one matching setup because generalizing formulation for more placeholder cases can be done as shown in Section A., as long as each type of item has sufficient placeholders for allocation. Therefore,
\[
|I_{s_p}|=|P_{s_p}|, \forall s_p \in S_p,
\]
and 
\[
|I|=|P|
\]
The assignment variable \( a_{ip} \) remains defined as before, indicating whether item \( i \) is assigned to placeholder \( p \).

\medskip
\noindent
\textbf{Type-Specific Assignment Constraints.}  
To ensure that items and placeholders are matched only within their respective types, we impose:
\[
\sum_{p \in P_{s_p}} a_{ip} = 1, \quad \forall i \in I_{s_p},
\qquad
\sum_{i \in I_{s_p}} a_{ip} = 1, \quad \forall p \in P_{s_p}, \quad \forall s_p \in S_p.
\]
These constraints enforce a one-to-one correspondence link from items to placeholders of the same type, ensuring that each item is assigned to exactly one compatible placeholder, and vice versa.

\textit{The routing structure follows the same principles as in the single-type formulation. }

\textbf{Remarks.}  
This generalized formulation accommodates multiple item types while preserving the assignment--routing coupling of the original model. Each type-specific subproblem maintains internal consistency within its type set, and the overall routing problem remains integrated through the joint decision variables \( x_{ij} \). Such an extension is particularly relevant for structured placement problems, including modular packing or multi-piece arrangement tasks (e.g., chessboard setup), where item--placeholder compatibility is type-dependent.
An application example for chess arrangement is provided in GitHub repository.

\paragraph{Use of Generative AI.}
This manuscript involved the use of generative AI tools (ChatGPT-4, OpenAI) during the drafting process. These tools were employed to assist with language refinement, equation formatting, and improving the clarity of technical explanations. All scientific content, methods, experimental results, and conclusions were developed and validated by the authors. Final manuscript versions were thoroughly reviewed to ensure accuracy and originality.

\bibliographystyle{unsrt}  
\bibliography{references}  

\end{document}